\documentclass[letterpaper]{article} 
\usepackage{aaai2026}  
\usepackage{times}  
\usepackage{helvet}  
\usepackage{courier}  
\usepackage[hyphens]{url}  
\usepackage{graphicx} 
\usepackage{natbib}  
\usepackage{caption} 
\usepackage{algorithm}
\usepackage{algorithmic}
\usepackage{amssymb}
\usepackage{amsmath}
\usepackage{booktabs}
\usepackage{multirow}
\usepackage{courier}
\usepackage{placeins}

\usepackage{newfloat}
\usepackage{listings}
\DeclareCaptionStyle{ruled}{labelfont=normalfont,labelsep=colon,strut=off} 
\floatstyle{ruled}
\newfloat{listing}{tb}{lst}{}
\floatname{listing}{Listing}
\title{DEIG: Detail-Enhanced Instance Generation with Fine-Grained Semantic Control}
\author{
    Shiyan Du\textsuperscript{\rm 1},
    Conghan Yue\textsuperscript{\rm 2,},
    Xinyu Cheng\textsuperscript{\rm 3,},
    Dongyu Zhang\textsuperscript{\rm 1,}\thanks{Corresponding author.}
}
\affiliations{
    \textsuperscript{\rm 1}Sun Yat-sen University, Guangzhou, China\\
    \textsuperscript{\rm 2}Fudan University, Shanghai, China\\
    \textsuperscript{\rm 3}Yale University, New Haven, United States\\
    dushy5@mail2.sysu.edu.cn, 
    chyue25@m.fudan.edu.cn,
    xinyu.cheng@yale.edu,
    zhangdy27@mail.sysu.edu.cn
}

\usepackage{bibentry}

\begin{document}

\maketitle

\begin{abstract}
Multi-Instance Generation has advanced significantly in spatial placement and attribute binding. However, existing approaches still face challenges in fine-grained semantic understanding, particularly when dealing with complex textual descriptions.
To overcome these limitations, we propose DEIG, a novel framework for fine-grained and controllable multi-instance generation. DEIG integrates an Instance Detail Extractor (IDE) that transforms text encoder embeddings into compact, instance-aware representations, and a Detail Fusion Module (DFM) that applies instance-based masked attention to prevent attribute leakage across instances. These components enable DEIG to generate visually coherent multi-instance scenes that precisely match rich, localized textual descriptions.
To support fine-grained supervision, we construct a high-quality dataset with detailed, compositional instance captions generated by VLMs. We also introduce DEIG-Bench, a new benchmark with region-level annotations and multi-attribute prompts for both humans and objects.
Experiments demonstrate that DEIG consistently outperforms existing approaches across multiple benchmarks in spatial consistency, semantic accuracy, and compositional generalization. Moreover, DEIG functions as a plug-and-play module, making it easily integrable into standard diffusion-based pipelines.
\end{abstract}

\begin{links}
    \link{Code}{https://github.com/dushy5/DEIG}
\end{links}

\section{Introduction}
Multi-Instance Generation \cite{li2023gligen, zheng2023layoutdiffusion, zhou2024migc, gu2025roictrl} has emerged as a promising direction in controllable image generation, where the goal is to generate images containing multiple semantically distinct instances at user-specified spatial locations. Recent approaches often build upon diffusion models, integrating spatial priors such as bounding boxes \cite{zhou2024migc}, masks \cite{kim2023dense, bar2023multidiffusion} or scribbles \cite{wang2024instancediffusion} to improve spatial alignment and identity consistency. These methods enable finer control over instance placement and visual composition, making them suitable for downstream applications such as fashion synthesis and artistic creation.

\begin{figure}[t]
\centering
\includegraphics[width=1.0\columnwidth]{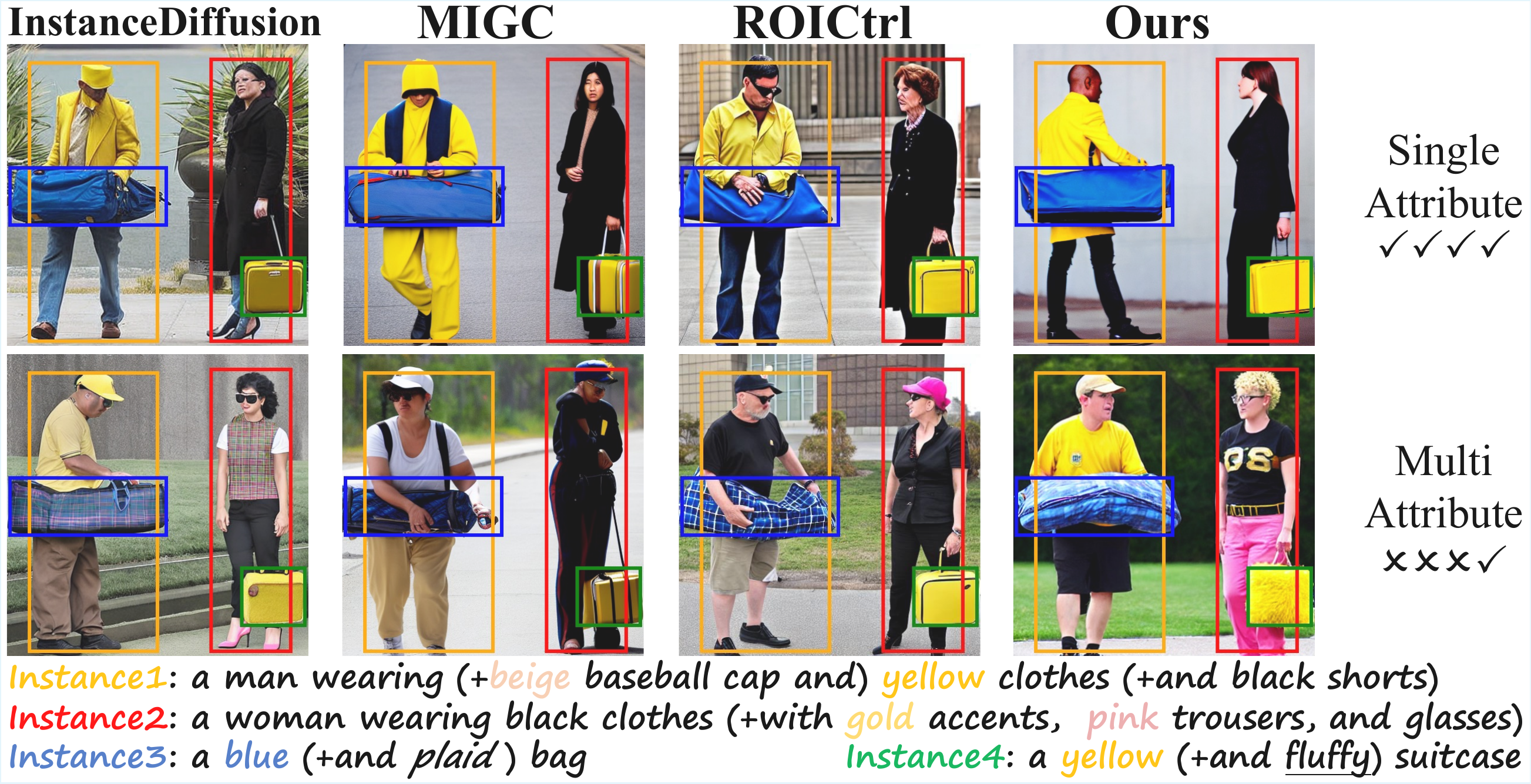}
\caption{\textbf{Fine-Grained Generation.}Given bounding boxes and detailed descriptions, our method accurately generate multi-attribute instances, while existing methods fail to preserve fine-grained semantic details.}
\label{fig_teaser}
\end{figure}

Nevertheless, current methods remain limited in their ability to handle rich and fine-grained region descriptions. As illustrated in Fig. \ref{fig_teaser}, although these methods can reliably generate instances with simple prompts, they often fail with complex, multi-attribute inputs involving multiple attributes such as multi-color designs and combinations of color, texture, and material. This restricts their applicability in scenarios requiring high-detail synthesis.

We attribute these limitations to two primary factors. First, current approaches predominantly focus on preventing semantic leakage, while neglecting the deeper semantic comprehension required for generating fine-grained visual details. \cite{zhou2024migc, wang2024instancediffusion, gu2025roictrl} Second, the training data used in these methods are typically annotated with coarse-grained templates, lacking detailed instance-level descriptions. This restricts the model’s ability to learn rich semantic-visual mappings from the data, thereby impeding the generation of visually and semantically coherent content.

In this work, we propose DEIG, a framework for generating Multi-Instance images with fine-grained attribute control. Inspired by recent advances in long-text alignment for text-to-image generation~\cite{hu2024ella, han2024emma}, we extend global prompt-based generation toward instance-level detail generation. Specifically, we introduce the \textit{Instance Detail Extractor} (IDE), which transforms high-dimensional embeddings from LLMs encoder into compact, instance-aware representations, enabling localized alignment between complex textual descriptions and visual regions. Leveraging a high-quality captioned dataset and a \textit{Detail Fusion Module} (DFM), DEIG enables precise and attribute-consistent multi-instance generation.

To evaluate instance-level controllability under fine-grained prompts, we introduce DEIG-Bench, a challenging benchmark specifically designed to address key limitations of existing datasets—namely, the underrepresentation of human instances and the reliance on single-attribute prompts. DEIG-Bench provides multi-attribute, compositional descriptions along with tailored evaluation protocols for both human and object instances. For human-centric scenes, we focus on color compositionality across wearable regions; for object-centric scenes, we gradually increase attribute complexity, encompassing color, material, and texture. To enable robust and reliable assessment, we leverage two distinct VLMs in a question-answering setup to evaluate complex semantic consistency.
To summarize, our main contributions are as follows:
\begin{itemize}
\item We propose DEIG, a novel framework that enhances instance-level detail representation and semantic understanding, to address the limitations of existing methods in handling rich, fine-grained region descriptions.
\item We introduce DEIG-Bench, a comprehensive evaluation suite specifically designed for multi-attribute, multi-Instance generation, to fill the gap of lacking benchmarks for evaluating fine-grained semantic prompts.
\item We conduct extensive experiments on multiple widely used benchmarks to demonstrate that DEIG significantly outperforms previous methods, particularly in generating detailed multi-attribute instances with improved semantic fidelity.
\end{itemize}

\section{Related Work}
\subsection{Controllable Diffusion models}
While text-to-image diffusion models generate high-quality outputs~\cite{ho2020denoising, rombach2022high, podell2023sdxl, betker2023improving}, recent work focuses on enhancing controllability by introducing various conditioning mechanisms. Subject-driven tasks~\cite{gal2022image, ruiz2023dreambooth, ma2024subject, wang2024ms, chen2024anydoor}, aim to preserve subject identity and enable personalized generation. Spatial control tasks~\cite{zhang2023adding, li2023gligen, mou2024t2i, mo2024freecontrol} specify the spatial arrangement of content using layout or structural information. Text-conditioned control targets fine-grained alignment with complex or compositional prompts~\cite{feng2022training, rassin2023linguistic, hu2024ella, han2024emma, liu2025llm4gen, wu2025paragraph}. Collectively, these tasks enhance the controllability of diffusion models and enable their application in different domains.

\subsection{Multi-Instance Generation}
With the widespread adoption of diffusion models, numerous studies have explored their potential for Multi-Instance Generation, focusing on synthesizing images with multiple semantically distinct instances arranged according to given layouts. Existing methods are typically categorized into training-free and training-based methods

\subsubsection{Training-Free Methods}
Training-free methods~\cite{kim2023dense, bar2023multidiffusion} primarily operate during inference by manipulating attention mechanisms~\cite{chen2024training, xiao2024r, phung2024grounded} or modifying loss functions~\cite{xie2023boxdiff, wang2025spotactor}. These methods often optimize the latent space to align generated instances with target spatial configurations. Although they offer considerable flexibility and require no retraining, they frequently compromise image fidelity, as the generation may deviate from the data distribution originally learned by the pretrained diffusion model.

\subsubsection{Training-Based Methods} 
In contrast, training-based methods incorporate instance-level cues, such as embeddings~\cite{li2023gligen, zhou2024migc, wang2024instancediffusion, gu2025roictrl} or semantic layouts~\cite{jia2024ssmg, wu2024ifadapter}, directly into the training pipeline. To mitigate attribute interference across instances, techniques like divide-and-conquer~\cite{zhou2024migc} and hierarchical generation~\cite{cheng2024hico} are employed. Recent DiT-based methods~\cite{zhang2024creatilayout, zhang2025eligen, zhou20253dis} achieve notable improvements in scalability and image quality. However, their high computational overhead makes them impractical for deployment on consumer-grade GPUs.

Despite their progress, existing methods struggle with compositional prompts involving fine-grained attributes, and remain limited in human-centric scenarios. This hinders their applicability in domains requiring precise semantic alignment and spatial control.

\begin{figure*}[t]
\centering
\includegraphics[width=1.0\textwidth]{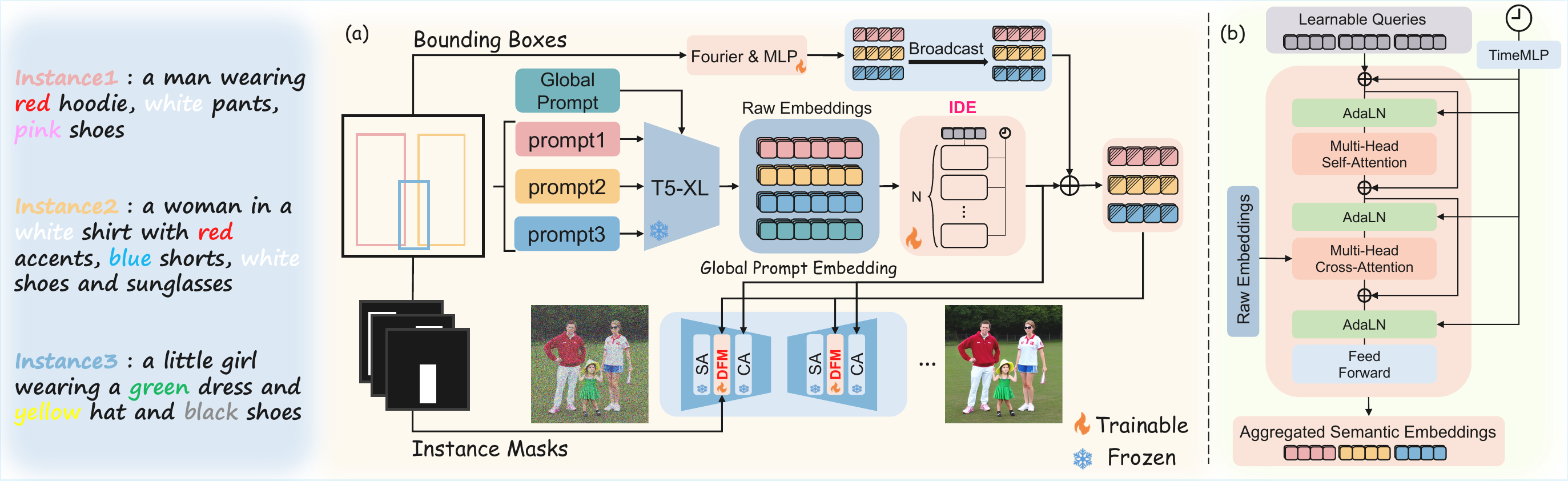}
\caption{\textbf{Overview of the DEIG pipeline.}
(a) DEIG enables the use of a frozen large text encoder to extract raw instance embeddings, which are refined by the IDE and fused into the UNet via the DFM for fine-grained control.
(b) Structure of the IDE, which refines learnable queries via time-aware self- and cross-attention to produce compact instance embeddings.}
\label{fig_pipe}
\end{figure*}

\section{Method}
\subsection{Overview}
In this task, users provide instance-level conditions, including bounding boxes, fine-grained descriptions, and a global context $\mathcal{P}$. Formally, the input is $\mathcal{C} = \{\mathcal{P}, (b_1, p_1), \dots, (b_n, p_n)\}$, where each $(b_i, p_i)$ denotes the location and semantics of the $i$-th instance. The objective is to generate an image that aligns with both spatial and semantic constraints while preserving global coherence. To this end, we propose DEIG, a pipeline for multi-instance generation with fine-grained semantic control over the attributes of each instance, as illustrated in Fig.~\ref{fig_pipe}.

\subsection{Instance-Level Semantic Enhancement}
As discussed above, existing methods often neglect instance-level semantic understanding, limiting their ability to generate fine-grained content. To address this, we introduce a dedicated module that effectively extracts and represents instance-specific semantics.

\subsubsection{Instance Detail Extractor}
Inspired by recent advances in long-text alignment with LLMs for text-to-image generation~\cite{hu2024ella, han2024emma, liu2025llm4gen, wu2025paragraph}, we replace traditional multi-modal encoders with a frozen text encoder to better capture fine-grained instance semantics. As the resulting embeddings are high-dimensional and costly to use directly, we introduce the \textit{Instance Detail Extractor} (IDE)—a lightweight module that distills rich text features into compact, instance-aware representations via learnable queries. While query-based distillation has proven effective in prior work~\cite{li2023blip, hu2024ella, han2024emma, he2024id, zhou2025fireedit}, our IDE is specifically designed for multi-instance generation, supporting token-level alignment and temporal conditioning compatible with diffusion models.

Structurally, IDE integrates stacked self-attention and cross-attention layers to refine instance-specific semantics, as illustrated in Fig.~\ref{fig_pipe}(b). It operates on encoded text features \( \mathbf{E}_{\tau} \in \mathbb{R}^{B \times N \times S_{\tau} \times C} \), where \( S_{\tau} \) is the embedding sequence length. To avoid direct use of these high-dimensional features, IDE introduces learnable queries \( \mathbf{Q} \in \mathbb{R}^{B \times N \times S \times C} \), with \( S \ll S_{\tau} \). We refer to \( S \) as the \textbf{Aggregated Semantic Dimension}, which acts as a bottleneck to compress and organize instance-specific information efficiently.

Each IDE layer refines the queries via temporally-aware attention. The process begins with timestep conditioning through a lightweight TimeMLP, followed by adaptive layer normalization (AdaLN)~\cite{perez2018film}, which adaptively modulates the query features using the same temporal embedding. A self-attention block captures intra-instance dependencies, and a subsequent cross-attention module aligns the queries with the high-dimensional text features from the frozen encoder. For the $i$-th layer, the key transformation can be expressed by:
\begin{equation}
\small
\begin{aligned}
\mathbf{H_{ca}^i} = \mathrm{CrossAttn}\left(\mathrm{AdaLN}\left(\mathbf{H_{sa}^i}, \mathbf{T_{emb}}\right), \left[\mathbf{H_{sa}^i}, \mathbf{E_{\tau}}\right]\right),
\end{aligned}
\end{equation}
where $\mathbf{T}_{\text{emb}}$ is the timestep embedding vector that modulates the visual features via AdaLN. $\mathbf{H_{sa}^i} \in \mathbb{R}^{B \times N \times S \times C}$ denotes the self-attended instance queries at the $i$-th layer, the output $\mathbf{H_{ca}^i} \in \mathbb{R}^{B \times N \times S \times C}$ captures cross-modal interactions between textual and visual embeddings, thereby completing the feature transformation at layer $i$.

\begin{figure*}[t]
\centering
\includegraphics[width=1.0\textwidth]{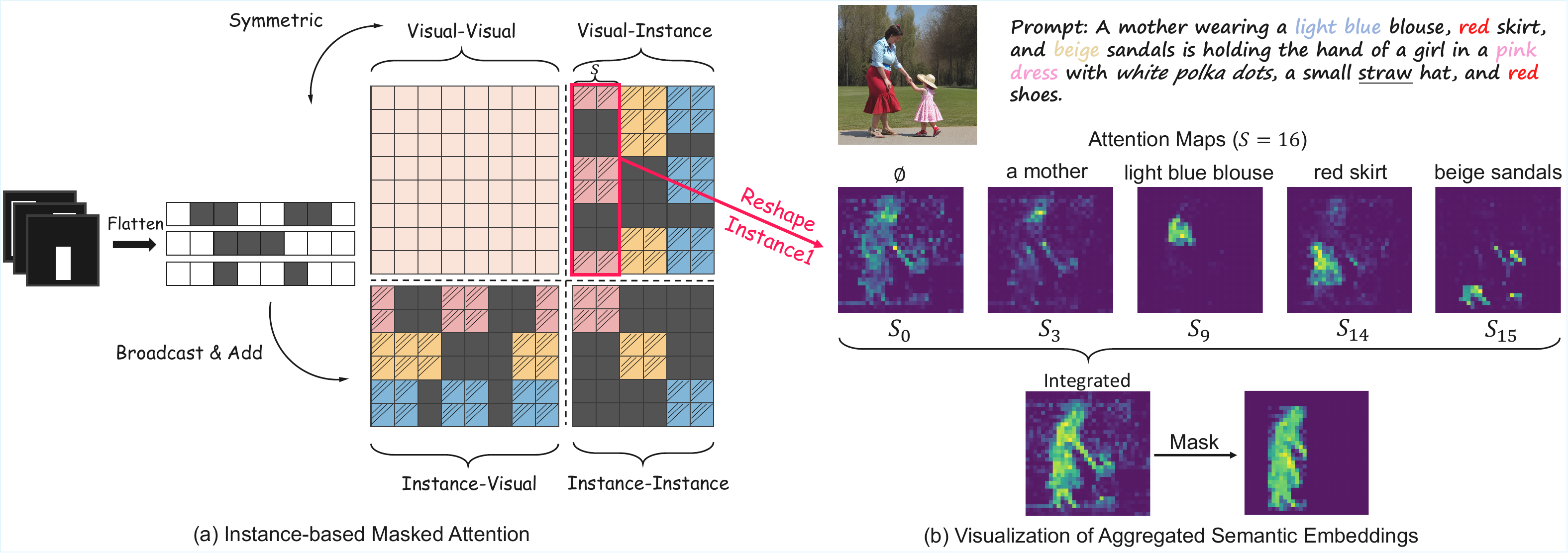} 
\caption{\textbf{Workflow Visualization of Fine-Grained Instance Generation.} (a) Instance-based masked attention mechanism divides the attention map into four sub-regions, applying masks to restrict cross-instance interactions and prevent semantic leakage. (b) Visualization of aggregated semantic embeddings across different semantic dimensions. }

\label{fig_attn}
\end{figure*}

The output is subsequently passed through a residual feed-forward network, completing one refinement cycle. After stacking $N$ such layers, IDE produces a set of compact and expressive instance-level embeddings, which we refer to as \textbf{Aggregated Semantic Embeddings}. 
We extract them from the second layer of the encoder in the UNet model at an intermediate diffusion timestep and visualize them across different semantic dimensions $S$. As illustrated in Fig.~\ref{fig_attn}(b), each dimension attends to specific fine-grained attributes described in the input. Collectively, these dimensions constitute a comprehensive representation of each instance, confirming that our approach achieves more precise semantic alignment than previous methods.

\subsection{Detail Fusion Module}
To integrate the Aggregated Semantic Embeddings into generation, we propose the \textit{Detail Fusion Module} (DFM), where Grounding Embeddings Broadcast aligns spatial cues with the semantic dimension, and Instance-based Masked Attention applies masking strategies to avoid attribute leakage.

\subsubsection{Grounding Embeddings Broadcast}
To align aggregated semantics with spatial generation, we adopt a broadcast-based spatial fusion approach across all \( S \) dimensions. Each instance’s spatial coordinate \( b_i \) is transformed via Fourier encoding and broadcast as:
\begin{equation}
\mathbf{f_{i}} = \mathcal{B}(\mathcal{F}(b_{i}), S), \quad \mathbf{e}_i = \mathcal{B}(\mathbf{e}_i, S) \in \mathbb{R}^{(B, N, S, C)}
\end{equation}
\begin{equation}
    \mathbf{G}_{\mathrm{ase}, i}=\mathrm{MLP}([m \cdot \mathbf{f_{i}} + (1-m)\cdot\mathbf{e}_i, \mathbf{E}_{\mathrm{ase}, i}])
\end{equation}
 
For $i$-th instance, \( \mathbf{f}_i \) denotes the broadcasted spatial embedding, and \( \mathbf{e}_i \) is a learnable null embedding used when spatial information is absent. \( \mathcal{F}(b_i) \) computes the Fourier encoding of coordinates, and \( \mathcal{B}(\cdot, S) \) denotes broadcasting along dimension \( S \). The binary mask \( m \in \{0,1\} \) selects between the two. The output $\mathbf{G}_{\mathrm{ase}, i}$ is a fused embedding, combining spatial and semantic cues, and is later used in instance-based masked attention.

\subsubsection{Instance-based Masked Attention}
Following prior embedding-guided fusion frameworks, we freeze the self- and cross-attention layers of UNet and insert a gated self-attention module between them to enable instance level semantic fusion. To mitigate attribute leakage across instances, which often occurs with standard self-attention, we introduce a masking mechanism based on instance partitioning.

As shown in Fig.~\ref{fig_attn}(a), the gated self-attention operates on the concatenation of visual and instance embeddings, generating an attention map naturally divided into four interpretable subregions. To regulate these interactions, we define a binary mask $\mathbf{M} \in \{0, -\infty\}^{L \times L}$, where $L = N_{\text{visual}} + N \times S$.
For any \( v \in \mathbf{V}_{\text{visual}} \), \( g \in \mathbf{G}_{\mathrm{ase}} \), the masking rules are as follows:

\begin{table*}[t]
\centering
\setlength{\tabcolsep}{0.7mm}{
\begin{tabular}{l|cccc|ccccc|cccc|ccccc|c}
\toprule
{} & \multicolumn{9}{c|}{\textbf{Qwen2.5-VL}}  & \multicolumn{9}{c|}{\textbf{InternVL3}} & {} \\  
\cmidrule(r){2-10} \cmidrule(r){11-19}
{\textbf{Method}} 
& \multicolumn{4}{c|}{$\textbf{MAA}_{\textbf{human}}$ $\uparrow$}
& \multicolumn{5}{c|}{$\textbf{MAA}_{\textbf{obj}}$ $\uparrow$}
& \multicolumn{4}{c|}{$\textbf{MAA}_{\textbf{human}}$ $\uparrow$}
& \multicolumn{5}{c|}{$\textbf{MAA}_{\textbf{obj}}$ $\uparrow$}
& {\textbf{mIoU} $\uparrow$} \\

\cmidrule(r){2-5} \cmidrule(l){6-10} 
\cmidrule(r){11-14} \cmidrule(l){15-19}
& C1 & C2 & C3 & AVG & L1 & L2 & L3 & L4 & AVG 
& C1 & C2 & C3 & AVG & L1 & L2 & L3 & L4 & AVG  \\
\midrule
GLIGEN
& 0.23 & 0.05 & 0.02 & 0.10 & 0.20 & 0.08 & 0.08 & 0.03 & 0.10
& 0.28 & 0.12 & 0.05 & 0.15 & 0.32 & 0.17 & 0.17 & 0.11 & 0.19
& 0.71 \\

MIGC
& 0.51 & 0.11 & 0.03 & 0.22 & 0.63 & 0.29 & 0.29 & 0.20 & 0.36
& 0.60 & 0.23 & 0.11 & 0.31 & 0.75 & 0.54 & 0.47 & 0.36 & 0.54
& 0.72 \\

InstanceDiffusion
& 0.54 & 0.17 & 0.05 & 0.25 & 0.53 & 0.24 & 0.32 & 0.20 & 0.33
& 0.61 & 0.23 & 0.12 & 0.32 & 0.62 & 0.49 & 0.46 & 0.36 & 0.49
& 0.75 \\

ROICtrl
& 0.61 & 0.23 & 0.09 & 0.31 & 0.56 & 0.32 & 0.27 & 0.16 & 0.33
& 0.68 & 0.33 & 0.16 & 0.39 & 0.67 & 0.48 & 0.42 & 0.30 & 0.47
& 0.71 \\

\midrule
\textbf{DEIG (Ours)} & \textbf{0.82} & \textbf{0.74} & \textbf{0.69} & \textbf{0.75} 
& \textbf{0.67} & \textbf{0.41} & \textbf{0.38} & \textbf{0.27} & \textbf{0.44}
& \textbf{0.86} & \textbf{0.82} & \textbf{0.81} & \textbf{0.83} 
& \textbf{0.79} & \textbf{0.58} & \textbf{0.50} & \textbf{0.44} & \textbf{0.58}
& \textbf{0.79}\\

\bottomrule
\end{tabular}
}
\caption{\textbf{Quantitative Results on DEIG-Bench.} We report \textit{MAA} for human (C1–C3) and object (L1–L4) instances under two VLMs, Qwen2.5-VL~\cite{bai2025qwen2} and InternVL3~\cite{zhu2025internvl3}, as well as spatial alignment via \textit{mIoU}.}

\label{tab:deigbench}
\end{table*}

\begin{table*}[t]
\centering
\setlength{\tabcolsep}{2.22mm}{
\begin{tabular}{l|cccccc|cccccc}
\toprule
\textbf{Method} & \multicolumn{6}{c|}{\textbf{Instance Success Rate (\%) $\uparrow$}} & \multicolumn{6}{c}{\textbf{mIoU} $\uparrow$ } \\
\midrule
Level & L2 & L3 & L4 & L5 & L6 & AVG & L2 & L3 & L4 & L5 & L6 & AVG \\
\midrule
GLIGEN & 41.56 & 32.29 & 28.13 & 25.38 & 29.79 & 29.91 & 36.70 & 29.10 & 24.92 & 23.37 & 27.22 & 27.03 \\
MIGC & 75.00 & 65.83 & 66.88 & 62.63 & 64.79 & 65.84 & 64.29 & 55.94 & 56.68 & 53.63 & 56.25 & 56.44 \\
InstanceDiffusion & 68.44 & 58.96 & 59.84 & 55.75 & 56.77 & 58.63 & 61.91 & 54.63 & 53.71 & 50.17 & 51.30 & 53.06 \\
ROICtrl & 72.19 & 65.63 & 64.69 & 59.88 & 60.94 & 63.25 & 61.75 & 57.32 & 56.28 & 52.68 & 53.56 & 55.27 \\
\midrule
\textbf{DEIG (Ours)} & \textbf{75.23} & \textbf{72.00} & \textbf{70.02} & \textbf{71.00} & \textbf{73.13} & \textbf{72.25} 
& \textbf{60.43} & \textbf{66.50} & \textbf{62.01} & \textbf{61.65} & \textbf{62.68} & \textbf{62.64} \\
\bottomrule
\end{tabular}
}
\caption{\textbf{Quantitative results on MIG-Bench.} MIG-Bench focuses on color-centric attribute evaluation via \textit{Instance Success Rate}, which checks color match within a predefined gamut, and assesses spatial alignment using \textit{mIoU} with Grounding-DINO.}

\label{tab:migbench}
\end{table*}

\begin{table}[t]
\centering
\setlength{\tabcolsep}{0.8mm}{
\begin{tabular}{l|cc|cc|ccc}
\toprule
\textbf{Method} & $\textbf{Acc}_{\textbf{c.}}$ & $\textbf{CLIP}_{\textbf{c.}}$ & $\textbf{Acc}_{\textbf{t.}}$ & $\textbf{CLIP}_{\textbf{t.}}$ & \textbf{AP} & $\textbf{AP}_{50}$ \\
\midrule
GLIGEN
&25.0 & 0.217 & 15.8 & 0.205 & 0.20 & 0.35\\
MIGC
&52.3 & 0.252 & 23.2 & 0.221 & 0.22 & 0.40\\
InstanceDiffusion
&53.4 & 0.252 & 25.9 & 0.227 & \textbf{0.40} & 0.57\\
ROICtrl
&56.9 & 0.255 & 23.7 & 0.223 & 0.26 & 0.51\\
\midrule
\textbf{DEIG (Ours)}
&\textbf{58.8} & \textbf{0.258} & \textbf{26.1} & \textbf{0.228} & 0.34 & \textbf{0.57} \\
\bottomrule
\end{tabular}
}
\caption{\textbf{Quantitative results on InstDiff-Bench.} Attribute alignment is assessed using \textit{Accuracy} and \textit{CLIP scores} for color ($\text{Acc}_{\text{c.}}$, $\text{CLIP}_{\text{c.}}$) and texture ($\text{Acc}_{\text{t.}}$, $\text{CLIP}_{\text{t.}}$), while spatial precision is measured by \textit{AP} and $\text{AP}_{50}$ using YOLOv8.}

\label{tab:instbench}
\end{table}

\noindent
\textbf{(a) Visual-Visual Attention.}
We observe that masking visual embeddings can noticeably degrade image fidelity. Therefore, we allow all visual embeddings to attend to each other without masking:
\begin{equation}
\mathbf{M}_{v_i, v_j} = 0 \quad, \forall i, j \in {1, \dots, N_{\text{visual}}}
\end{equation}
\textbf{(b) Symmetric Instance–Visual Attention. }
Each instance embedding is allowed to attend only to visual embeddings from the same instance, and vice versa. For any two different instances, interactions between them are masked by setting the corresponding attention scores a value of negative infinity. This operation can be formulated as:
\begin{equation}
\small
\mathbf{M}_{v_i, g_j} = \mathbf{M}_{g_i, v_j} = 
\begin{cases}
0, & \text{if }\text{Instance}(v_i) = \text{Instance}(g_j) \\
-\infty, & \text{otherwise}
\end{cases}
\end{equation}
\textbf{(c) Instance-Instance Attention.}
Instance embeddings attend only to others within the same semantic group and all cross-group interactions are masked with negative infinity, following the same rule as above.
\begin{equation}
\mathbf{M}_{g_i, g_j} =
\begin{cases}
0, & \text{if } \text{Group}(g_i) = \text{Group}(g_j) \\
-\infty, & \text{otherwise}
\end{cases}
\end{equation}
The final masked attention output is computed via a standard self-attention:
\begin{equation}
    \hat{\mathbf{A}} = \mathrm{Softmax}\left( \frac{\mathbf{Q} \mathbf{K}^T}{\sqrt{d}} + \mathbf{M} \right) \mathbf{V}
    \label{eq_attn}
\end{equation}

As shown in Fig.~\ref{fig_attn}(b), the mask effectively suppresses attention that would otherwise leak into other instances or the background.
Finally, the visual embeddings are then updated through gated residual mechanism:
\begin{equation}
\mathbf{V}_{visual} = \mathbf{V}_{visual} + \eta \cdot \tanh\gamma \cdot \mathcal{ES}( \hat{\mathbf{A}}),
\end{equation}
where $\mathcal{ES}(\cdot)$ denotes \textit{embedding slices} operation, which extract a subset of the visual embedding outputs. $\eta$ and $\gamma$ are learnable scalars that control the update strength.

\begin{figure}[t]
\centering
\includegraphics[width=1.0\columnwidth]{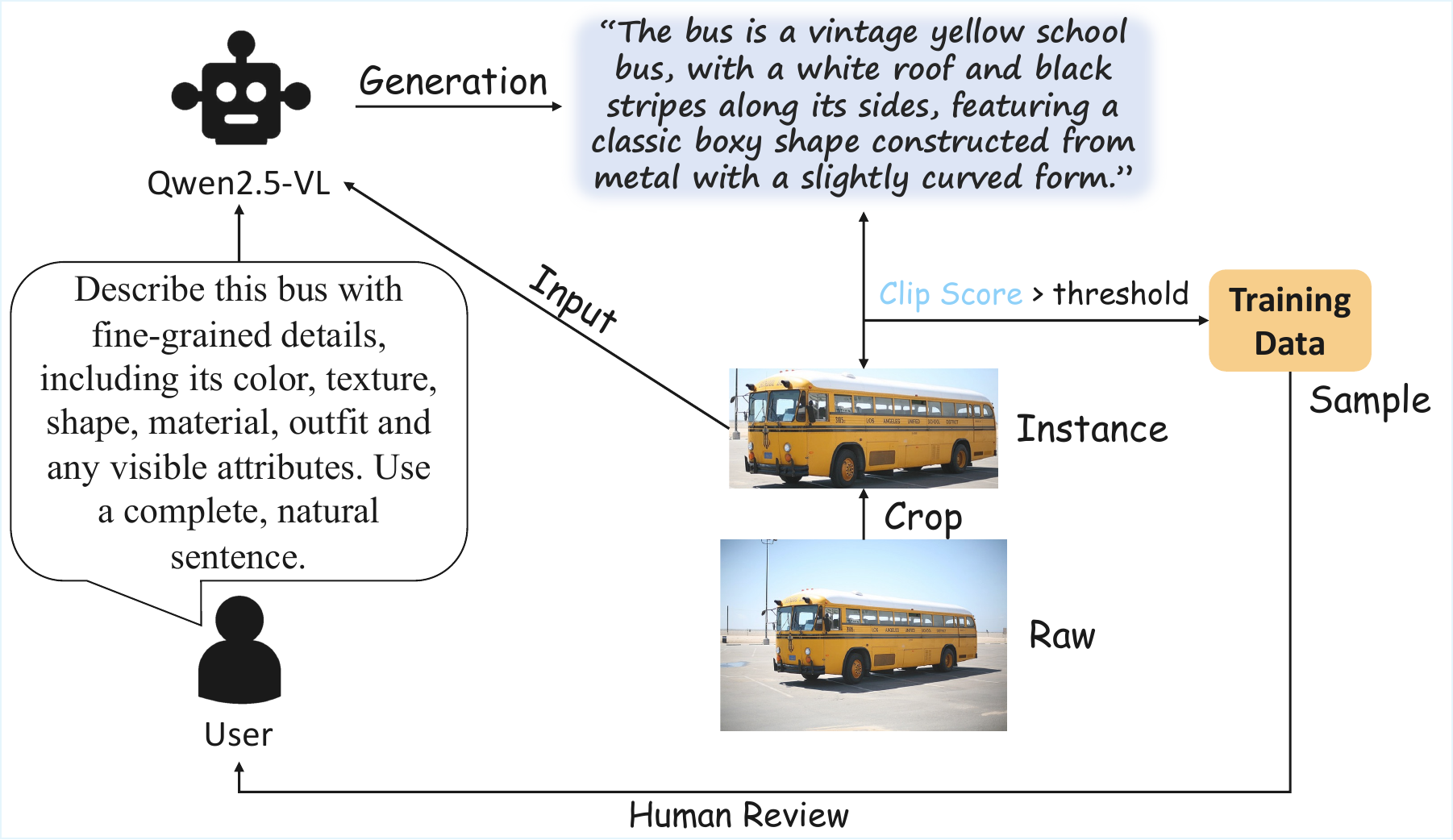}
\caption{\textbf{Detail-Enriched caption generation pipeline.} Instances are described by a VLM from cropped images and filtered using CLIP scores and human review.}
\label{fig_qwen}
\end{figure}

\subsection{Detail-Enriched Instance Captions}
High-quality datasets are vital for accurate instance-level generation. We curate a dataset from MS-COCO~\cite{lin2014microsoft} using Qwen2.5-VL~\cite{bai2025qwen2} to generate detailed, context-aware captions averaging 20–30 words per instance. Unlike template-based methods, our approach produces natural and semantically rich descriptions. To ensure quality and minimize hallucinations, we remove grayscale and low-fidelity images via VLM assessment and apply a two-stage verification process. This process involves (1) computing the CLIP~\cite{radford2021learning} score for each image–caption pair and keeping those above a predefined threshold, and (2) performing human verification on a random subset of 500 pairs to confirm overall consistency and reliability. The data construction pipeline is shown in Fig.~\ref{fig_qwen}.

\begin{figure*}[ht]
\centering
\includegraphics[width=1.0\textwidth]{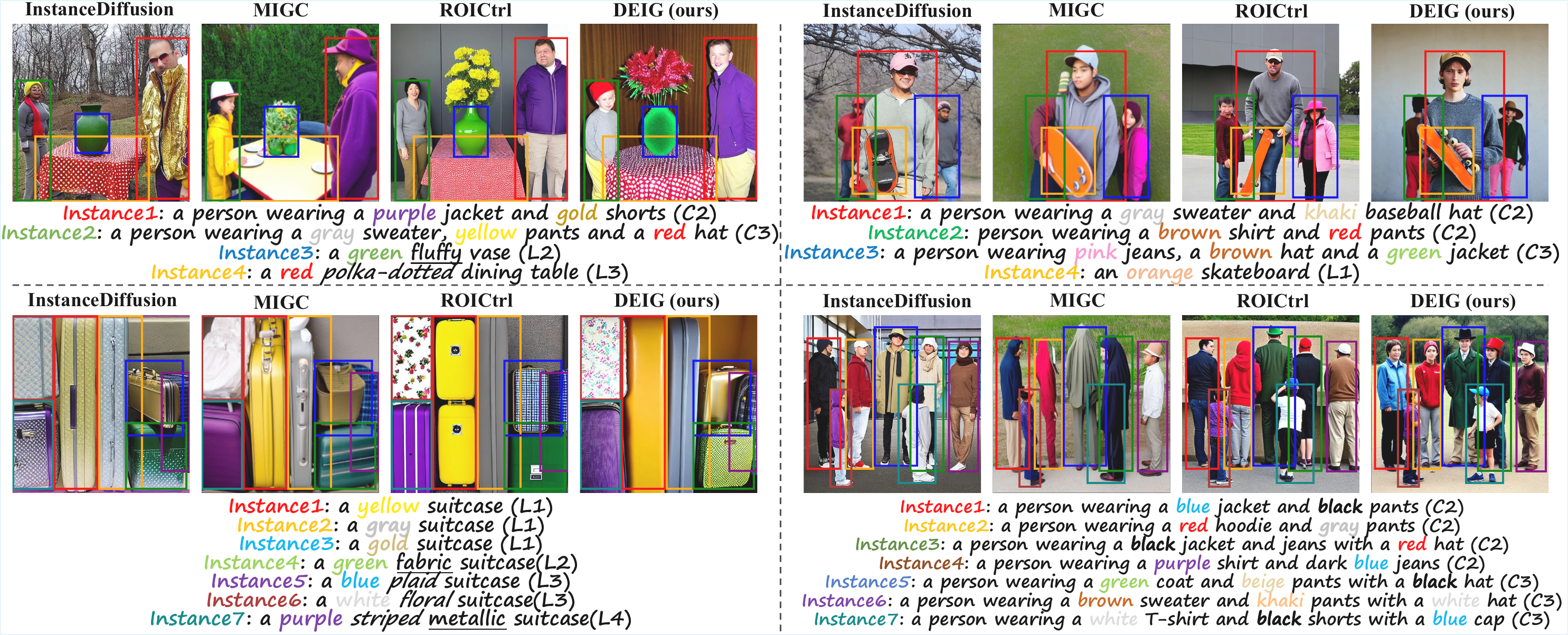}
\caption{\textbf{Qualitative comparison on DEIG-Bench.} DEIG exhibits accurate generation of fine-grained, multi-attribute instances across varying levels of complexity, demonstrating superior compositional control and semantic alignment.}
\label{fig_qual}
\end{figure*}

\section{Experiment}
To thoroughly evaluate the effectiveness of DEIG, we conduct comprehensive experiments, including comparisons with previous state-of-the-art baselines and detailed ablation studies. We adopt the encoder of Flan-T5-XL~\cite{chung2024scaling} as the text encoder in all experiments. Additional experimental settings and more qualitative results are provided in the Appendix.

\subsection{DEIG-Bench}
Existing Multi-Instance Generation benchmarks often lack fine-grained supervision and realistic human-centric scenes. Descriptions like "a red person" fail to capture real-world appearance complexity. Moreover, prior datasets typically focus on single-attribute prompts, limiting the evaluation of compositional understanding.

To address these gaps, we introduce \textbf{DEIG-Bench}, a benchmark for evaluating fine-grained, multi-attribute generation of both human and object instances. It features compositional prompts and structured complexity levels that reflect real-world attribute entanglement. DEIG-Bench is constructed from 400 filtered images in the MS-COCO validation set, ensuring each image contains 3–10 visible instances for reliable attribute recognition.

DEIG-Bench evaluates fine-grained compositionality in both human and object instances. For humans, we define three difficulty levels: C1, C2 and C3, based on color combinations across wearable regions. For objects, we define four attribute levels from L1 to L4, where L1 includes only color, L2 adds material on top of color, L3 adds texture on top of color, and L4 combines all three for the highest semantic and visual complexity. Evaluation combines \textit{mIoU}, computed using Grounding-DINO~\cite{liu2024grounding}, for spatial alignment, and two VLMs for semantic validation. We also introduce \textit{Multi-Attribute Accuracy} (MAA) to measure how well models bind multiple attributes to each instance, which is calculated as the ratio of instances correctly identified by the VLM to the total number of instances.

For broader validation, we additionally report results on MIG-Bench~\cite{zhou2024migc} and InstDiff-Bench~\cite{wang2024instancediffusion}.

\subsection{Comparison}
We compare our method with previous SOTA approaches for Multi-Instance Generation. Specifically, we select GLIGEN, MIGC, InstanceDiffusion, and ROICtrl as baselines for comparison.

\begin{table}[t]
\centering
\setlength{\tabcolsep}{1.2mm}
\begin{tabular}{ccc|ccc}
\toprule
\textbf{IDE} & \textbf{DFM} & \textbf{Cap.} & \textbf{mIoU} $\uparrow$ & $\textbf{MAA}_{\textbf{human}}$ $\uparrow$ & $\textbf{MAA}_{\textbf{obj}}$ $\uparrow$ \\
\midrule
           & \checkmark & \checkmark & 0.73  & 0.51 & 0.35 \\
\checkmark &            & \checkmark & 0.75  & 0.70  & 0.41 \\
\checkmark & \checkmark &            & 0.70  & 0.31  & 0.29 \\
\midrule
\checkmark & \checkmark & \checkmark & \textbf{0.79} & \textbf{0.75} & \textbf{0.44} \\
\bottomrule
\end{tabular}
\caption{\textbf{Ablation study of DEIG components.} We evaluate the individual contributions of IDE, DFM, and Captions supervision by measuring MAA with Qwen2.5-VL and mIoU.}

\label{tab:ablation}
\end{table}

\subsubsection{Quantitative Results}
Tab.~\ref{tab:deigbench} presents the quantitative results on DEIG-Bench. On human-centric tasks, DEIG significantly outperforms all baselines across attribute complexities, particularly under multi-color combinations, indicating stronger compositional generalization enabled by our detail-aware semantic extraction and fusion modules. For object-centric generation, DEIG also achieves consistently higher scores, with notable gains on color attributes. In contrast, improvements on material and texture are more modest, likely because color correlates more directly with the RGB space and is easier for diffusion models to learn, whereas material and texture require more abstract semantic understanding beyond surface-level appearance.

\begin{figure}[t]
\centering
\includegraphics[width=1.0\columnwidth]{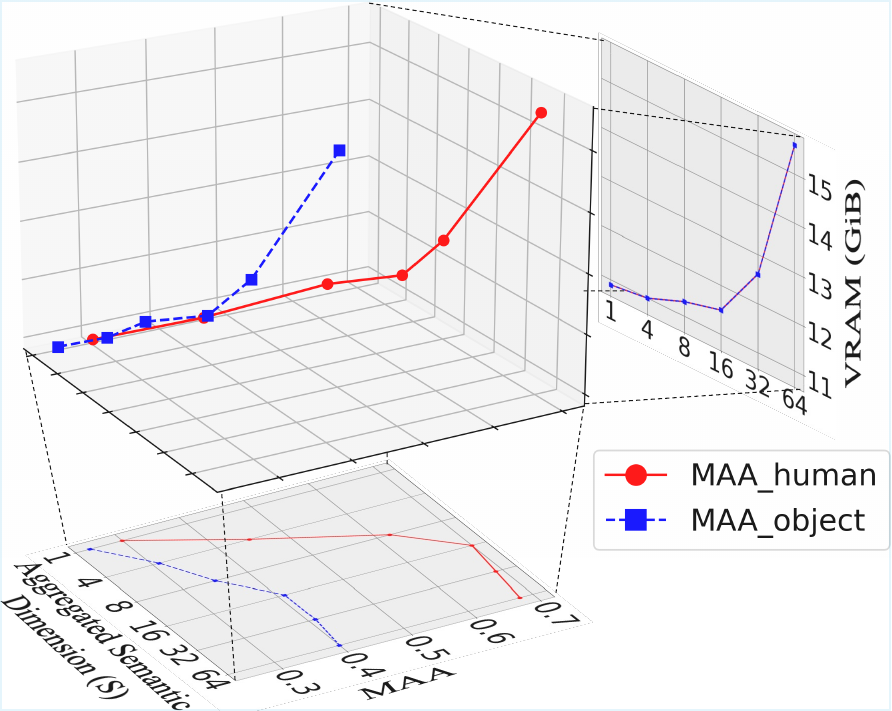}
\caption{\textbf{Trade-off Between Semantic Precision and Computational Cost.} 3D visualization of how the aggregated semantic dimension \( S \) affects MAA and GPU memory usage under FP16 precision.}
\label{fig_ablation}
\end{figure}

\begin{figure*}[t]
\centering
\includegraphics[width=1.0\textwidth]{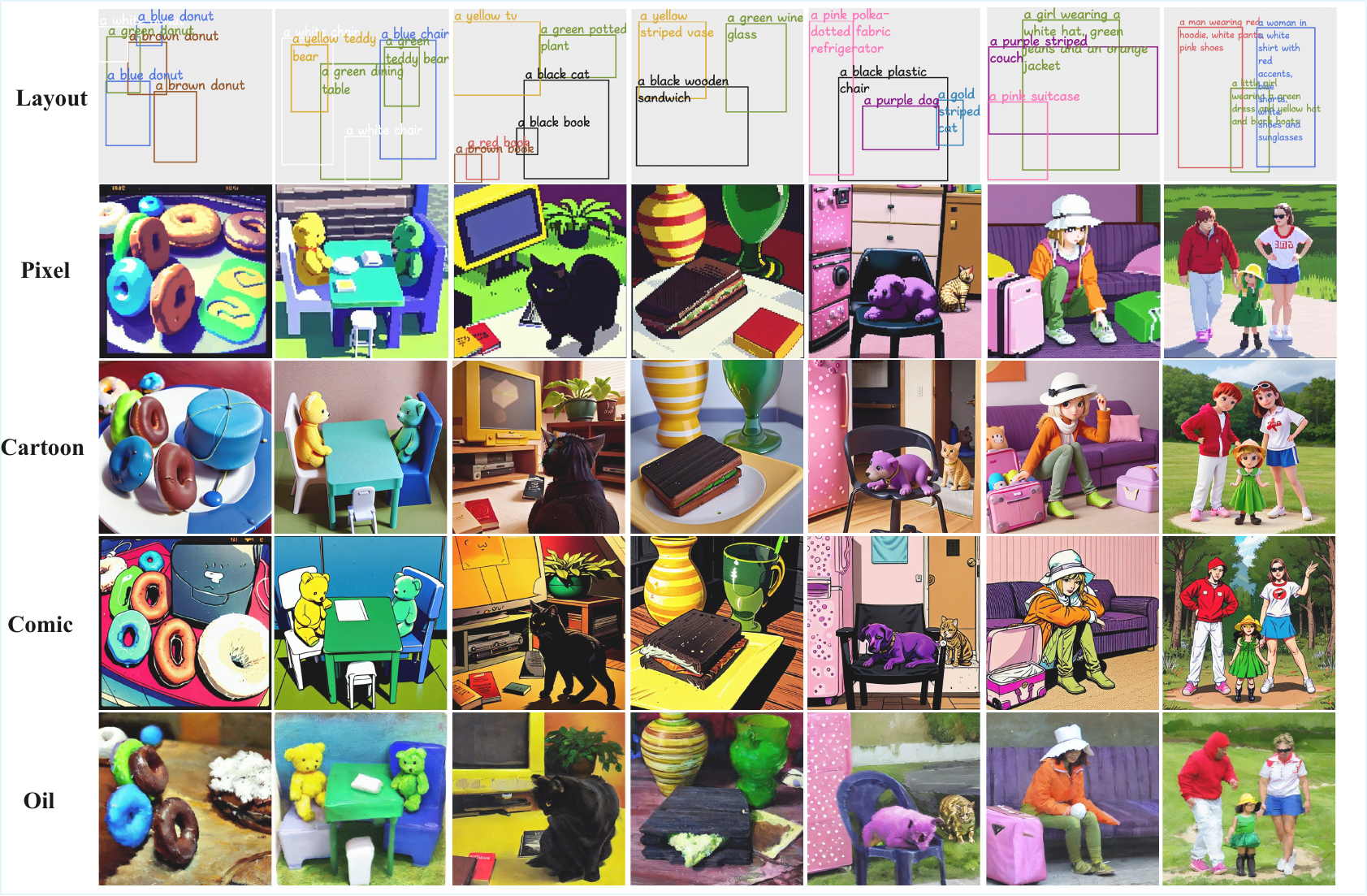}
\caption{\textbf{Plug-and-play adaptation to community diffusion models.} DEIG preserves fine-grained semantic control when integrated into community diffusion models, demonstrating strong compatibility and generalization.}
\label{fig_community}
\end{figure*}

Tab.~\ref{tab:migbench} and Tab.~\ref{tab:instbench} present the results on MIG-Bench and InstDiff-Bench. DEIG shows strong controllability under dense layouts and diverse attribute prompts. While spatial alignment on InstDiff-Bench is slightly lower due to instance-masked attention limiting interactions in crowded regions, it consistently surpasses baselines in accuracy and CLIP alignment, especially for color attributes.

\subsubsection{Qualitative Comparison}
DEIG effectively generates images from multi-attribute prompts with distinct spatial separation and strong attribute fidelity, as illustrated in Fig.~\ref{fig_qual}. While other methods struggle to represent multiple attributes details per instance, our approach preserves semantic integrity across instances, with each accurately reflecting its described color, material, and texture combination.

Fig.~\ref{fig_community} shows DEIG adapted to a community diffusion backbone without retraining. It preserves spatial layout and fine-grained generation quality, highlighting the plug-and-play nature and compatibility with common diffusion pipelines.

\subsection{Ablation Study}
We conduct an ablation study to assess the contributions of three components in DEIG, including IDE, DFM, and detail-enriched instance captions. As shown in Tab.~\ref{tab:ablation}, removing the captioning module causes the largest drop in MAA, highlighting the importance of fine-grained semantic supervision. Excluding IDE decreases semantic alignment, confirming its role in detailed instance representation, while removing DFM slightly reduces accuracy due to semantic leakage between instances. Overall, the accuracy drop for object instances is smaller than for human instances, indicating that human-centric generation is more sensitive to fine-grained control loss.

We further study the effect of the aggregated semantic dimension \( S \) on model performance and computational cost. As shown in Fig.~\ref{fig_ablation}, increasing \( S \) improves MAA for both human and object instances, with gains saturating around \( S = 16 \). Beyond this, performance plateaus or slightly declines due to overfitting. Meanwhile, GPU memory usage rises steadily with \( S \), as seen in the side view, indicating a trade-off between precision and efficiency. A setting of \( S = 16 \sim 32 \) provides a good balance.

\section{Conclusion}
In this paper, we propose DEIG, a detail-enhanced framework for fine-grained multi-instance generation. DEIG integrates instance-aware semantic extraction and masked attention fusion to improve attribute alignment and reduce leakage. Built as a plug-and-play module, it adapts to existing diffusion pipelines with minimal overhead. Extensive experiments on multiple benchmarks show that DEIG achieves strong spatial consistency and semantic fidelity, advancing controllable generation in complex multi-instance scenarios.

\section{Acknowledgments}
This research is supported by the Guangdong Basic and Applied Basic Research Foundation under Grant No. 2024A1515011741. The authors express their sincere gratitude to all collaborators and colleagues who provided valuable insights, constructive suggestions throughout the course of this work.

\bibliography{aaai2026}

\appendix
\clearpage
\begin{huge}
    \noindent\textbf{Appendix}
\end{huge}

\setcounter{figure}{0}
\renewcommand{\thefigure}{A\arabic{figure}}

\section*{Detailed Experiment Settings}

\subsection*{Implementation Details}
We train and evaluate models at a resolution of 512×512, initializing from the pre-trained GLIGEN checkpoint based on Stable Diffusion v1.4. Training is conducted for 800k iterations using 8 NVIDIA RTX 3090 GPUs. We adopt the AdamW optimizer with a constant learning rate of 1e-4 and apply a linear warm-up over the first 10k iterations. The batch size is set to 4 with gradient accumulation of 4, resulting in an effective batch size of 128. For parameter settings, we set the aggregated semantic dimension to $S = 16$, and the number of IDE layers to $N = 6$.

\subsection*{DEIG-Bench Construction Details} 
\begin{figure}[!htbp]
\centering
\includegraphics[width=1.0\columnwidth]{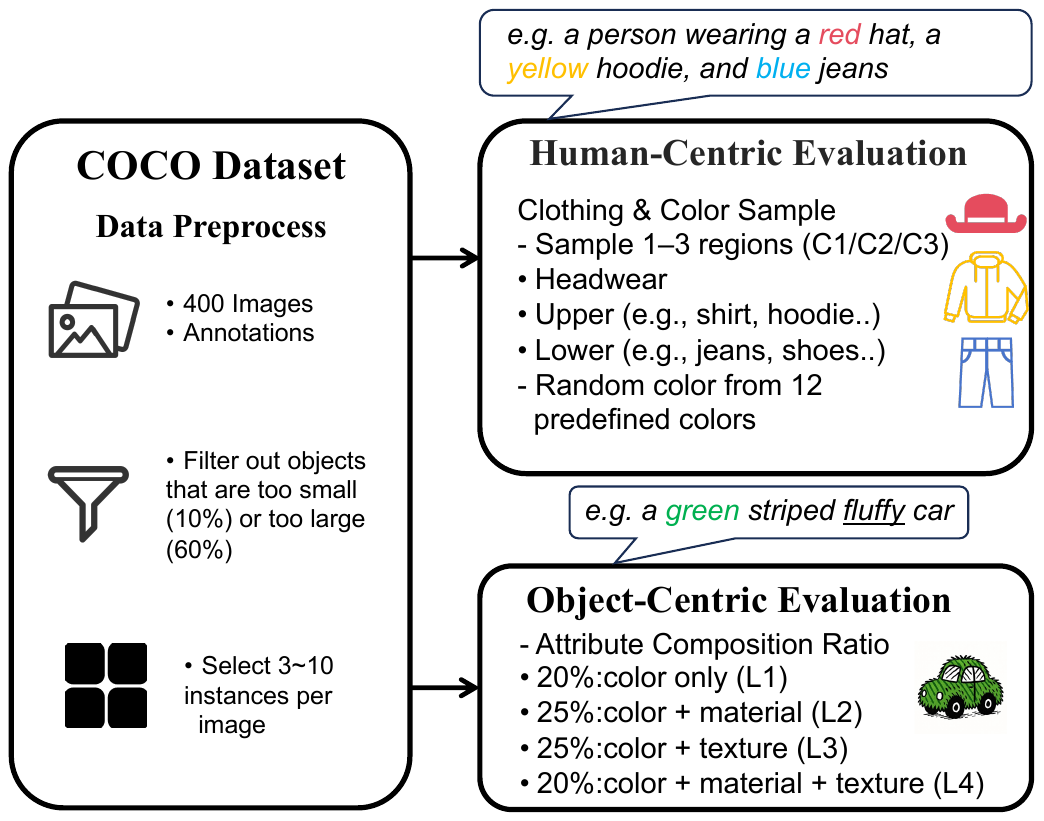} 
\caption{\textbf{The pipeline of constructing DEIG-Bench.}}
\label{data}
\end{figure}

To evaluate our methods on fine-grained attribute understanding, we constructed deig-bench based on the MSCOCO validation set, focusing on three key capabilities: clothing color recognition for persons, and material and texture or pattern recognition for general objects. The construction process is illustrated in Fig.~\ref{data}. We first filtered images, selecting only those containing at least one annotated instance. For each image, object instances were retained if their relative area with respect to image size was between 10\% and 60\%. This filtering ensured that each selected instance was sufficiently visible for attribute recognition while avoiding extremely large objects that might dominate the scene or obscure other content.
We further restricted each image to a maximum of 10 and a minimum of 3 qualifying instances. A total of 400 images were randomly sampled and successfully processed. For each selected instance, synthetic textual descriptions were generated according to category-specific strategies.

To evaluate the model’s ability to generate fine-grained, out-of-distribution (OOD) instance-level descriptions, we construct a benchmark that systematically incorporates compositional attributes. Specifically, we define a vocabulary of 13 common colors (e.g., \textit{red}, \textit{blue}, \textit{green}, \textit{yellow}, \textit{black}, \textit{white}, \textit{gray}, \textit{brown}, \textit{orange}, \textit{purple}, \textit{pink}, \textit{gold}), 8 canonical materials (\textit{rubber}, \textit{fluffy}, \textit{metallic}, \textit{wooden}, \textit{plastic}, \textit{fabric}, \textit{leather}, \textit{glass}), and 4 representative textures or patterns (\textit{striped}, \textit{plaid}, \textit{floral}, \textit{polka-dotted}).

For person categories, we randomly select one to three clothing regions (i.e., hat, upper body, and lower body), and assign each a sampled color and garment type. The resulting descriptions follow a natural template, such as \texttt{a person wearing a red hat and blue pants}.

For object categories, instance descriptions are generated following a probabilistic attribute composition strategy: 30\% of the samples use only color, 25\% combine color and material, 25\% combine color and texture or pattern, and the remaining 20\% include all three attributes. This produces diverse and compositional phrases such as \texttt{a green plastic bottle} or \texttt{a red striped fabric pillow}.

Finally, we construct the global prompt for each image by concatenating all instance-level descriptions with commas, forming a unified input sequence that reflects the full semantic content of the scene.

\subsection*{MIG-Bench} 
MIG-Bench is a benchmark designed to evaluate spatial alignment and regional attribute grounding. It consists of 800 images randomly sampled from the MSCOCO validation set, where each image is annotated with uniquely colored instances. Captions are generated using the coarse template \texttt{a-[colored]-[noun]}, such as \texttt{a red car}, to enforce explicit color grounding.

During construction, small-sized regions and overly dense images are filtered out to ensure clarity and reduce visual ambiguity. For evaluation, MIG-Bench employs \textit{Grounding-DINO} to detect the referenced objects, and compares the predicted bounding boxes and corresponding color gamut against predefined thresholds. Performance is reported using two metrics: \textit{mIoU} to measure spatial alignment, and \textit{Instance Success Rate} to assess whether the generated captions successfully align both location and described color attributes for each instance.

\subsection*{InstDiff-Bench} 
InstDiff-Bench is designed to evaluate multi-instance generation across diverse spatial input formats and attribute-level controls, based on the MSCOCO validation set. 

In our setting, bounding boxes are used as the primary input modality for specifying instance regions. For spatial alignment evaluation, YOLOv8 is employed to compute standard object detection metrics such as \textit{Average Precision} (AP), primarily on in-distribution instance captions.
InstDiff-Bench defines a pool of 8 common colors and 8 common textures. For each instance, a single adjective (either a color or a texture) is randomly sampled from this pool, and instance-level captions are constructed using the template \texttt{[adj.]-[noun]} (e.g., \texttt{blue person}, \texttt{metallic lamp}). The cropped region corresponding to 

\clearpage

\begin{figure*}[!htbp]
\centering
\includegraphics[width=0.85\textwidth]{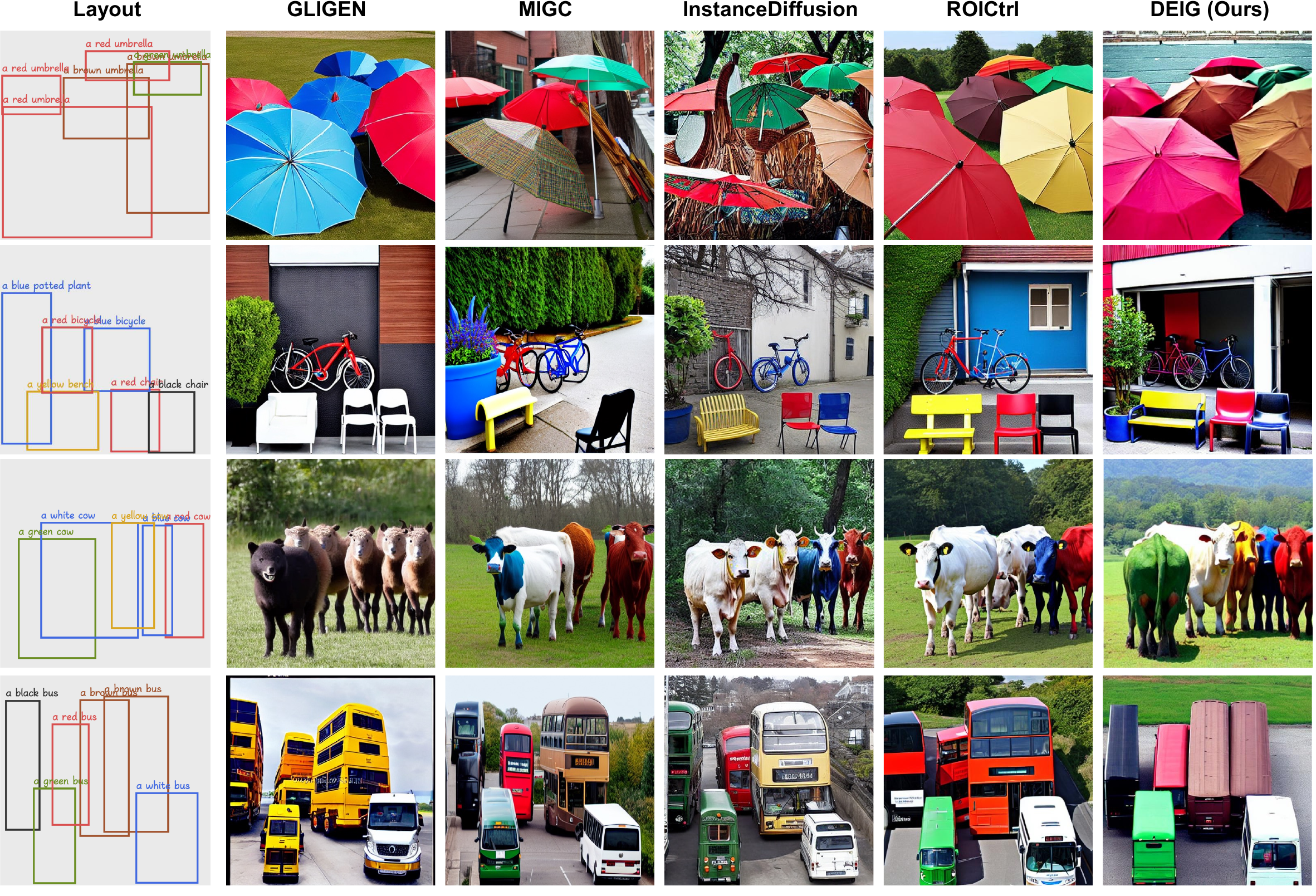} 
\caption{\textbf{More qualitative results on MIG-Bench.}
MIG-Bench focuses on spatial layout and color consistency.}
\label{mig}
\end{figure*}

\begin{figure*}[!htbp]
\centering
\includegraphics[width=0.85\textwidth]{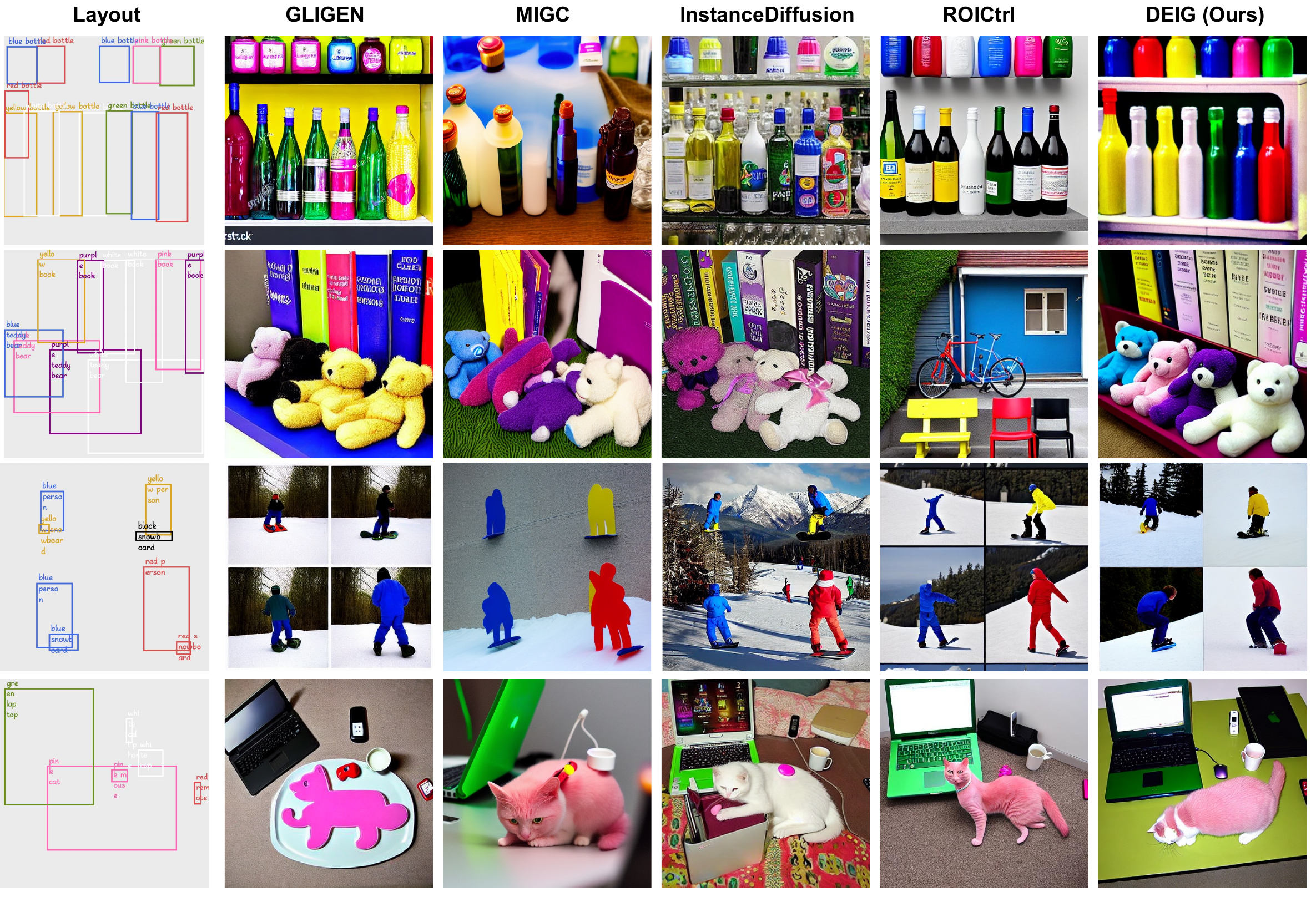} 
\caption{\textbf{More qualitative results on InstDiff-Bench.}
InstDiff-Bench imposes strict spatial constraints and contains a larger number of instances per image. }

\label{instdiff}
\end{figure*}

\begin{figure*}[!htbp]
\centering
\includegraphics[width=0.95\textwidth]{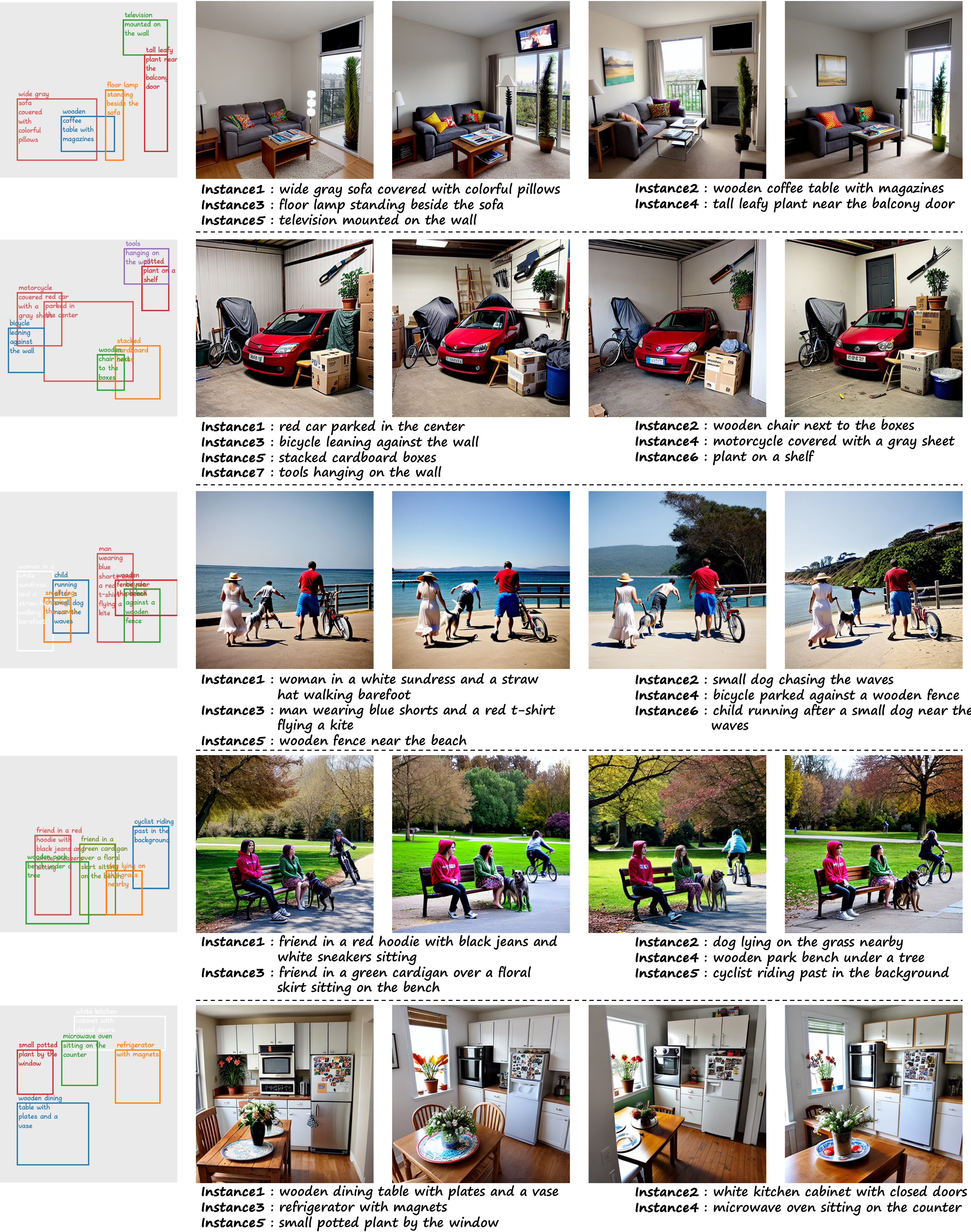} 
\caption{\textbf{More qualitative results using Free-Form generation.}
The examples illustrate that our model effectively handles open-ended descriptions across diverse scenes, producing visually coherent outputs with fine-grained instance attributes and spatial relationships without relying on predefined templates.}

\label{freeform}
\end{figure*}
\FloatBarrier

\noindent
each bounding box is fed into a CLIP model, which predicts visual attributes. The benchmark then evaluates whether the predicted color or texture aligns with the ground-truth description using attribute-level accuracy metrics (Acc${\text{color}}$, Acc${\text{texture}}$), and also reports the regional CLIP similarity score for each instance.

\section*{More Qualitative Results}
We provide additional qualitative comparisons and results to illustrate the effectiveness and diversity of our methods. Fig.\ref{mig}, and Fig.\ref{instdiff} show qualitative examples from \textbf{MIG-Bench}, and \textbf{InstDiff-Bench}, respectively.

In addition, we provide qualitative examples of \textbf{Free-Form generation} without template constraints, as shown in Fig.\ref{freeform}. 
Since our model is trained on non-template image–text pairs, it naturally adapts to open-ended descriptions and produces coherent, visually aligned outputs even in unconstrained settings.

\section*{Limitations}

\begin{figure}[!htbp]
\centering
\includegraphics[width=1.0\columnwidth]{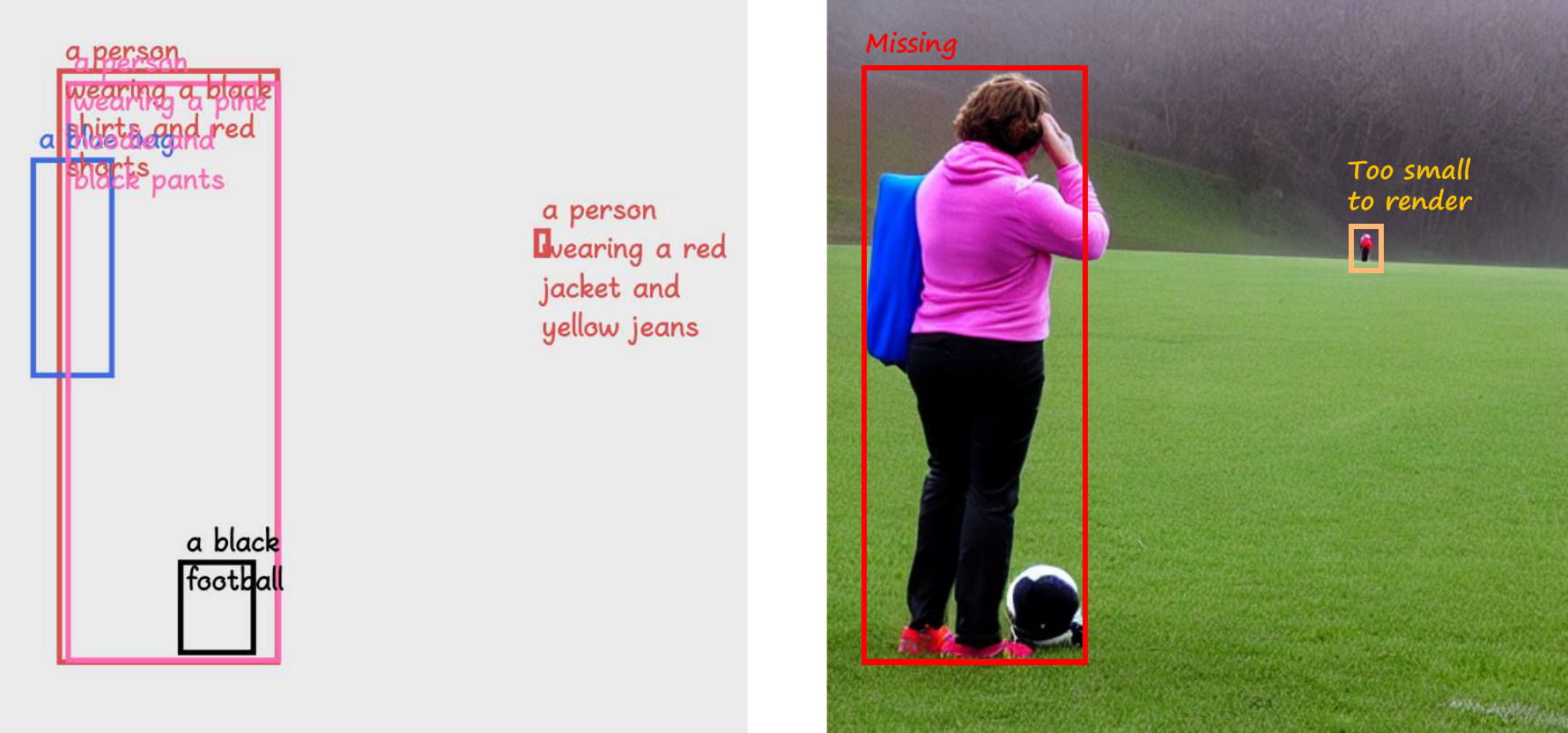} 
\caption{\textbf{Illustration of typical limitations.}
Examples show cases where overlapping objects, dense scenes, or small instances cause incomplete generation or loss of fine-grained details.}
\label{limi}
\end{figure}
\FloatBarrier

Despite the strong performance of our approach, several limitations remain. When the scene contains a large number of objects with significant spatial overlap, as observed in benchmarks such as InstDiff-Bench, the model may fail to disentangle adjacent instances, resulting in incomplete or unstable generations, as illustrated in Fig.~\ref{limi}. This challenge becomes more severe in densely populated scenes where complex spatial relationships exist. Moreover, the model has difficulty preserving fine-grained details for small objects, where limited spatial resolution hinders accurate attribute depiction. Future work will focus on improving the handling of overlapping regions and enhancing the model’s ability to render detailed attributes in cluttered environments.

\section*{Societal Impacts}
Our work introduces an instance-level controllable text-to-image generation framework that enhances fine-grained visual details while supporting precise spatial layout and attribute manipulation. Such capability has the potential to significantly benefit creative industries by improving the efficiency and consistency of content production in domains such as animation, design, advertising, and education. By lowering the technical barriers to producing high-quality visuals, this technology democratizes the creative process and enables a wider range of users to participate in content creation.

At the same time, the enhanced controllability of generative models also raises important societal concerns. The ability to synthesize highly specific and realistic images could be misused for generating misleading or harmful content, posing risks to privacy, public trust, and information integrity. Furthermore, as automated generation tools become more powerful, they may disrupt traditional creative workflows and lead to economic challenges for professionals reliant on manual methods. Finally, biases present in training datasets may be amplified through model outputs, reinforcing undesirable stereotypes. Addressing these risks requires not only technical safeguards, such as content filtering and usage guidelines, but also responsible deployment practices and continued attention to ethical considerations in generative model research.

\end{document}